\documentclass[10pt,twocolumn,letterpaper]{article}

\usepackage{iccv}
\usepackage{times}
\usepackage{epsfig}
\usepackage{graphicx}
\usepackage{amsmath}
\usepackage{amssymb}

\usepackage{booktabs}
\usepackage[utf8]{inputenc} % allow utf-8 input
\usepackage[T1]{fontenc}    % use 8-bit T1 fonts
\usepackage{url}            % simple URL typesetting
\usepackage{booktabs}       % professional-quality tables
\usepackage{amsfonts}       % blackboard math symbols
\usepackage{nicefrac}       % compact symbols for 1/2, etc.
\usepackage{microtype}      % microtypography
\usepackage{xcolor}         % colors
\usepackage{bm}         % colors
\usepackage{amsmath}         % colors
\usepackage{graphicx}

\usepackage{multirow}
\usepackage{makecell}
\usepackage{pifont}
\usepackage{wrapfig}

\usepackage{caption}
\usepackage{subcaption}
\usepackage{graphicx}
\usepackage{comment}
\usepackage{makecell}

\usepackage[breaklinks=true,bookmarks=false]{hyperref}

\iccvfinalcopy % *** Uncomment this line for the final submission

\def\iccvPaperID{5538} % *** Enter the ICCV Paper ID here

\ificcvfinal\fi

\newcommand{\nickname}{SHERF}

\begin{document}

%%%%%%%%% TITLE
\title{\nickname{}: Generalizable Human NeRF from a Single Image}

\author{
Shoukang Hu\textsuperscript{\rm 1}\footnotemark[1] \quad
Fangzhou Hong\textsuperscript{\rm 1}\footnotemark[1] \quad
Liang Pan\textsuperscript{\rm 1} \quad
Haiyi Mei\textsuperscript{\rm 2} \quad
Lei Yang\textsuperscript{\rm 2} \quad
Ziwei Liu\textsuperscript{\rm 1}\\
\textsuperscript{\rm 1}S-Lab, Nanyang Technological University \quad
\textsuperscript{\rm 2}SenseTime Research\\
}

% \author{First Author\\
% Institution1\\
% Institution1 address\\
% {\tt\small firstauthor@i1.org}
% % For a paper whose authors are all at the same institution,
% % omit the following lines up until the closing ``}''.
% % Additional authors and addresses can be added with ``\and'',
% % just like the second author.
% % To save space, use either the email address or home page, not both
% \and
% Second Author\\
% Institution2\\
% First line of institution2 address\\
% {\tt\small secondauthor@i2.org}
% }

\twocolumn[{
    \renewcommand\twocolumn[1][]{#1}%
    \maketitle
    \vspace{-20pt}
    \setlength{\abovecaptionskip}{0cm}
    \begin{center}
        \centering
        \includegraphics[width=1.0\textwidth]{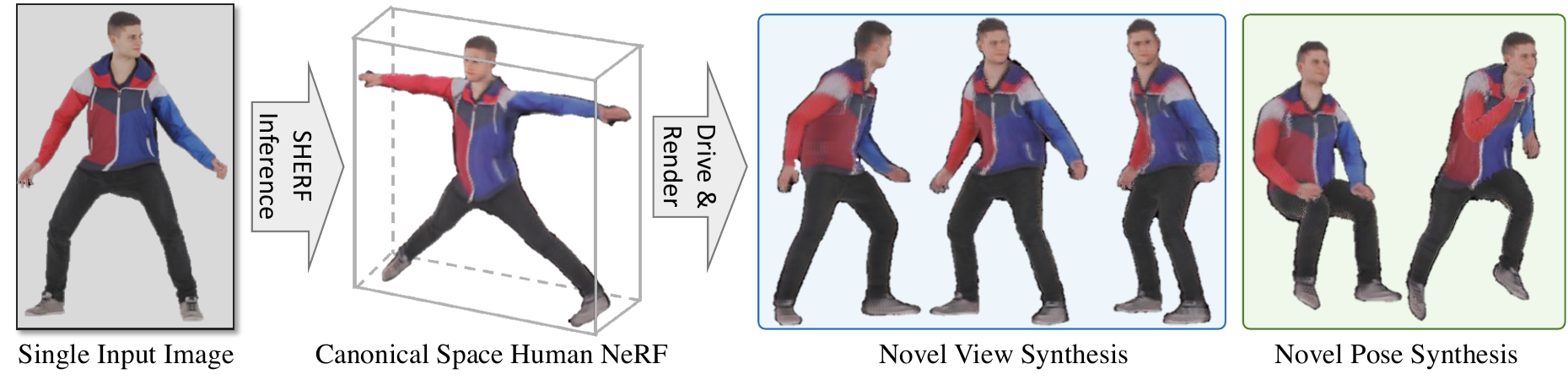}
        \captionof{figure}{\textbf{\nickname{} is a single image-based generalizable Human NeRF.} With just one inference pass on a single image, \nickname{} reconstructs Human NeRF in the canonical space which can be driven and rendered for novel view and pose synthesis.}
        \label{fig:teaser}
    \end{center}
    % \vspace{-6pt}
}]

\renewcommand{\thefootnote}{\fnsymbol{footnote}}
\footnotetext[1]{Equal contribution}

\maketitle
% Remove page # from the first page of camera-ready.
\ificcvfinal\thispagestyle{empty}\fi

%%%%%%%%% ABSTRACT

\begin{abstract}

Existing Human NeRF methods for reconstructing 3D humans typically rely on multiple 2D images from multi-view cameras or monocular videos captured from fixed camera views. However, in real-world scenarios, human images are often captured from random camera angles, presenting challenges for high-quality 3D human reconstruction.
In this paper, we propose \textbf{\nickname{}}, the first \textbf{generalizable Human NeRF model} for recovering animatable 3D humans from a single input image. \nickname{} extracts and encodes 3D human representations in canonical space, enabling rendering and animation from free views and poses.
To achieve high-fidelity novel view and pose synthesis, the encoded 3D human representations should capture both global appearance and local fine-grained textures. To this end, we propose a bank of 3D-aware hierarchical features, including global, point-level, and pixel-aligned features, to facilitate informative encoding.
Global features enhance the information extracted from the single input image and complement the information missing from the partial 2D observation. Point-level features provide strong clues of 3D human structure, while pixel-aligned features preserve more fine-grained details.
To effectively integrate the 3D-aware hierarchical feature bank, we design a feature fusion transformer.
Extensive experiments on THuman, RenderPeople, ZJU\_MoCap, and HuMMan datasets demonstrate that \nickname{} achieves state-of-the-art performance, with better generalizability for novel view and pose synthesis.
Our code is available at \url{https://github.com/skhu101/SHERF}.

\end{abstract}

%%%%%%%%% BODY TEXT
\section{Introduction}
\label{sec:intro}

Human NeRFs aim to recover high-quality 3D humans from 2D observations, avoiding the need to capture ground truth 3D geometry information~\cite{peng2021neural,chen2021animatable, animatablenerf, xu2021h, noguchi2021neural, weng2022humannerf, su2021nerf, jiang2022selfrecon, jiang2022neuman, wang2022arah, kwon2021neural, gao2022mps, choi2022mononhr, zhao2021humannerf}.
The development of Human NeRFs addresses a long-standing scientific request and has the potential to enable real-world applications \eg VR/AR. By leveraging Human NeRF, we can reconstruct 3D humans directly from 2D observations, saving time and effort to collect ground truth 3D information.

Existing Human NeRF methods can be classified into two categories. 
The first category focuses on reconstructing 3D humans from monocular or multi-view videos~\cite{peng2021neural,chen2021animatable, animatablenerf, xu2021h, noguchi2021neural, weng2022humannerf, su2021nerf, jiang2022selfrecon, jiang2022neuman, wang2022arah}.
These methods optimize subject-specific Human NeRF, which are time-consuming and not suitable for the rapid applications of Human NeRF.
To address the slow optimization process, the second category of Human NeRF methods~\cite{kwon2021neural, gao2022mps, choi2022mononhr, zhao2021humannerf} propose to learn generalizable Human NeRF models. These methods can reconstruct Human NeRF from a few multi-view human images in a single forward pass, which largely speeds-up the process.
Although these methods can achieve acceptable performance in 3D human reconstruction, they require multi-view images under well-defined camera angles, limiting their applicability in real-world scenarios where only a single image with a random angle is available.
MonoNHR~\cite{choi2022mononhr} addresses this gap by exploring novel view synthesis from a single image. But it cannot animate the reconstructed Human NeRF with novel poses, still limiting its applicability.

Recovering animatable 3D humans from a single human image with generalizable Human NeRF is a challenging problem due to two main challenges.
The first challenge is the \textbf{missing information} from the partial observation.
Existing generalizable Human NeRF~\cite{kwon2021neural, gao2022mps} focus too much on local feature preservation, while struggle to complement the missing information.
The second challenge is reconstructing \textbf{animatable} 3D humans from a single human image.
To make animatable Human NeRF from partial observations, it is necessary to complete missing appearance while also ensuring coherent understanding of 3D human structure. This poses additional challenges beyond the task of simply completing missing information.

In this work, we propose \textbf{\nickname{}}, the first generalizable Human NeRF based on single image inputs.
We propose a hierarchical feature bank to address the challenge of information missing from the single image input. This feature bank includes global, point-level, and pixel-aligned features, which enable informative 3D human representations encoding. The hierarchical feature bank captures both the global human structure and local fine details, which are essential for high-fidelity human NeRF reconstruction.
In addition, we introduce a feature fusion transformer to effectively merge features in the hierarchical feature bank. As illustrated in Fig.~\ref{fig:teaser}, our method can reconstruct correct colors for visible areas and provide plausible guesses for non-observable areas. The former is attributed to the fine-grained 3D-aware features that are crucial for reconstructing accurate geometry and color details, while the latter is enabled by the global features that allow color inference of invisible parts. The combination of these abilities leads to our method's capability of generating high-quality novel views and poses.

To address the challenge of animatability, \nickname{} models the 3D human representation in canonical space, making it amenable to pose transformation and rendering. We use the SMPL prior~\cite{SMPL:2015} to transform hierarchical features extracted from the input image to the canonical space, where they are encoded to better complete missing information and acquire the human structure information.

We evaluate \nickname{} on several datasets including THuman~\cite{tao2021function4d}, RenderPeople~\cite{renderpeople}, ZJU\_MoCap~\cite{neuralbody} and HuMMan~\cite{cai2022humman}.
Our results show that \nickname{} outperforms previous state-of-the-art generalizable Human NeRF methods in both novel view and novel pose synthesis with single images as inputs.
We also conduct a detailed analysis on the effects of varying input camera views, which provides further insights into \nickname{}.
Our main contributions are as follows:

\noindent \textbf{1)} To the best of our knowledge, \nickname{} is the first generalizable Human NeRF model to recover animatable 3D humans from a single human image. It pushes the boundaries of Human NeRF to a more general setting and bridges the gap of applying Human NeRF in real-world scenarios.

\noindent \textbf{2)} With 3D-aware hierarchical features, \nickname{} learns both fine-grained and global features to recover texture details and complement information missing from partial observations.

\noindent \textbf{3)} \nickname{} achieves state-of-the-art performance compared with previous generalizable Human NeRF methods~\cite{kwon2021neural, gao2022mps} in both novel view and novel pose synthesis on four large-scale datasets.
%-------------------------------------------------------------------------
\section{Related Work}
\label{sec:related_work}

\noindent \textbf{Human NeRF.}
NeRF~\cite{nerf,advancesinnerf} has inspired research in 3D human reconstruction. Human NeRF can synthesize high-fidelity novel views or poses of 3D humans, given multi-view or monocular human videos. Neural Body~\cite{peng2021neural} applies sparse convolutions to model the radiance volume, while others model human NeRF in the canonical space~\cite{chen2021animatable, animatablenerf, xu2021h, noguchi2021neural, weng2022humannerf, su2021nerf, jiang2022selfrecon, jiang2022neuman, wang2022arah} using SMPL LBS weights or optimizing LBS weights with appearance.
While these methods achieve impressive results, they often require time-consuming optimization and dense observations. To address this, there has been a growing interest in generalizable human NeRF~\cite{kwon2021neural, gao2022mps, choi2022mononhr, zhao2021humannerf, huang2022elicit}. These methods require fewer observations and only one forward pass.
This work also aims to develop generalizable human NeRF and tackle a more challenging scenario, recovering animatable human NeRF from a single image.

\noindent \textbf{Monocular Human Reconstruction.}
Statistical 3D human models~\cite{SMPL:2015, smplx, joo2018total, romero2022embodied, xu2020ghum} have enabled the reconstruction of 3D humans from monocular observations. Using these models, researchers have estimated coarse human shapes and poses~\cite{kanazawa2018end, kocabas2020vibe, kocabas2021pare, kolotouros2019learning}. To model the complex shape of clothed humans, mesh deformation is estimated~\cite{alldieck2018detailed, alldieck2018video, alldieck2019tex2shape, pons2017clothcap, zhu2019detailed}. Implicit representations, such as SDF, have been used to improve geometry quality~\cite{saito2019pifu, saito2020pifuhd, he2020geo, li2020robust, dong2022pina, li2020monocular, bozic2021neural, yang2021s3}. To take advantage of both explicit and implicit representations, researchers have explored combining these representations~\cite{bhatnagar2020combining, bhatnagar2020loopreg, huang2020arch, he2021arch++, zheng2021pamir, xiu2022icon, xiu2022econ, alldieck2022photorealistic, corona2022structured} for better generalizability and reconstruction quality. In comparison, with the advantages of NeRF, we do not need 3D ground truth for training. Moreover, we reconstruct humans in the canonical space, which can be easily driven with novel poses.

\begin{figure*}[t]
    \centering
    \includegraphics[width=16cm]{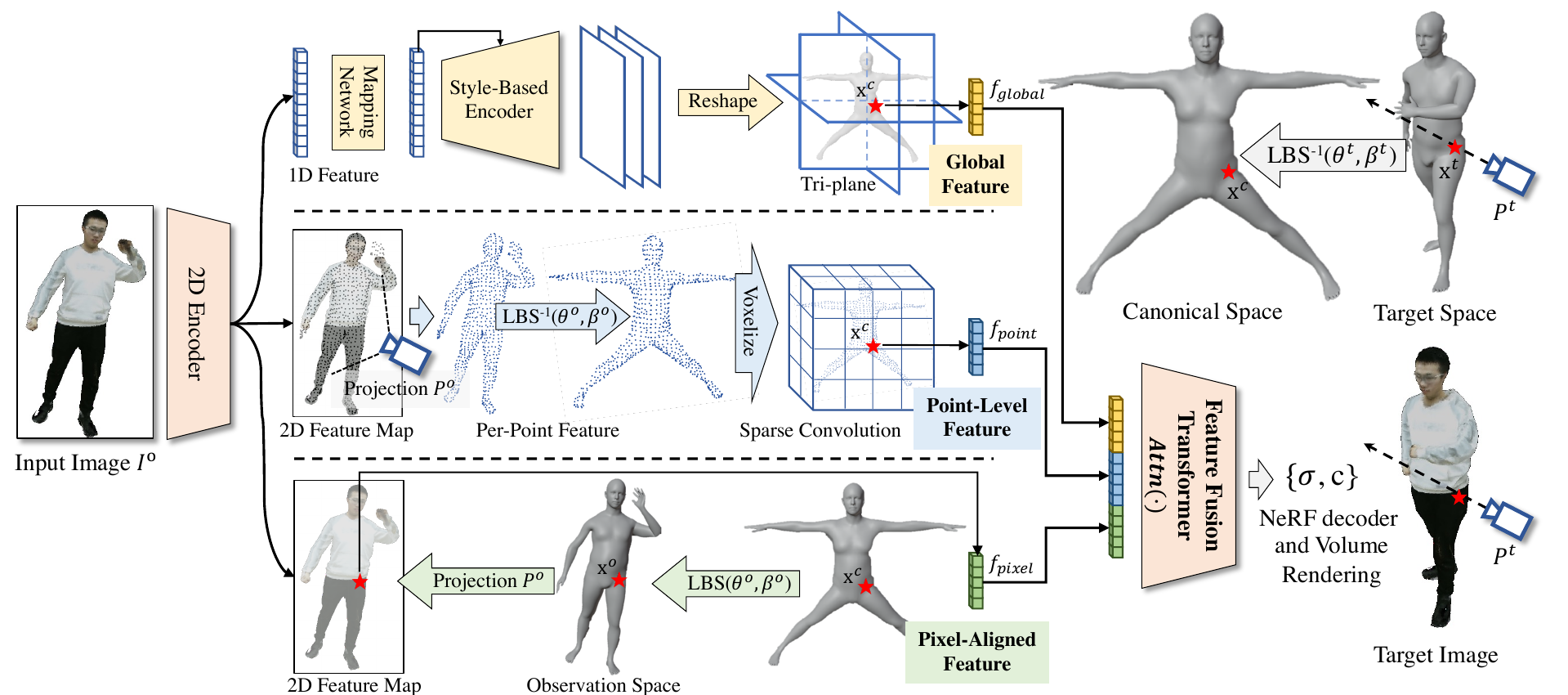}
    \caption{\textbf{\nickname{} Framework.} To render the target image, we first cast rays and sample points in the target space. The sample points are transformed to the canonical space through inverse LBS. We then query the corresponding 3D-aware global, point-level, and pixel-aligned features. The deformed points, combined with the bank of features, are input into the feature fusion transformer and NeRF decoder to get the RGB and density, which are further used to produce the target image through volume rendering.} 
\label{fig: overview}
% \vspace{-3mm}
\end{figure*}

\noindent \textbf{Generalizable NeRF.}
% cross-scene multi-view aggregator
% single view
NeRF requires dense calibrated views~\cite{nerf, muller2022instant}, but recent advances have led to the development of generalizable NeRF that can work with very few or even single views. Cross-scene multi-view aggregators~\cite{wang2021ibrnet, chen2021mvsnerf, liu2022neuray, wang2022generalizable} can synthesize novel views by learning to aggregate sparse views. Other works~\cite{gao2020portrait,lin2022vision,sitzmann2019scene,mi2022im2nerf,guo2022fast,sajjadi2022scene,cao2022fwd} encode observations to latent space and decode to NeRF.
Our work focuses on generalizable Human NeRF that encodes single human images into the canonical 3D space.
\section{Our Approach}
\label{sec:method}

\subsection{Preliminary}

\noindent \textbf{NeRF}~\cite{mildenhall2020nerf} learns an implicit continuous function which takes as input the 3D location $\mathbf{x}$ and viewing direction $\mathbf{d}$ of each point and predicts the volume density $\mathbf{\sigma} \in [0, \infty)$ and color value $\mathbf{c}\in [0,1]^3$, \ie, $F_{\Phi}: (\gamma(\mathbf{x}), \gamma(\mathbf{d})) \to (\mathbf{c}, \mathbf{\sigma})$, where $F_{\Phi}$ is parameterized by a multi-layer perception (MLP) network, $\gamma$ is the positional embedding.
To render the RGB color of pixels in the target view, rays are cast from the camera origin $\mathbf{o}$ through the pixel with the direction $\mathbf{d}$. 
Based on the classical volume rendering~\cite{kajiya1984ray}, the expected color $\hat{C}(\mathbf{r})$ of the camera ray $\mathbf{r}(t) = \mathbf{o} + t\mathbf{d}$ is computed as 
\begin{equation}
\label{eqn:volume_rendering}
\begin{aligned}
\hat{C}(\mathbf{r})=\int_{t_n}^{t_f} T(t) \sigma(\mathbf{r}(t)) \mathbf{c}(\mathbf{r}(t), \mathbf{d})dt,
\end{aligned}
\end{equation}
where $t_n$ and $t_f$ denote the near and far bounds, $T(t)=\exp(-\int_{t_n}^{t}\sigma(\mathbf{r}(s))ds)$ denotes the accumulated transmittance along the direction $\mathbf{d}$ from $t_n$ to $t$.
In practice, the continuous integral is approximated with the quadrature rule~\cite{max1995optical} and reduced to the traditional alpha compositing.

\noindent \textbf{SMPL}~\cite{SMPL:2015} is a parametric human model 
% $M(\bm{\beta}, \bm{\theta})$ 
which defines $\bm{\beta}, \bm{\theta}$ to control body shapes and poses. 
In this work, we apply the Linear Blend Skinning (LBS) algorithm of SMPL to transform points from the canonical space to target/observation spaces. 
Formally, a 3D point $\bm{x}^c$ in the canonical space is transformed to an observation space defined by pose $\bm{\theta}$ as
% \begin{equation} \label{eq:lbs}
$\bm{x}^o = \sum_{k=1}^{K}w_{k}\bm{G_k}(\bm{\theta}, \bm{J})\bm{x}^c$,
% \end{equation}
where $K$ is the joint number, $\bm{G_k}(\bm{\theta}, \bm{J})$ is the transformation matrix of joint $k$, $w_{k}$ is the blend weight. 
The transformation from target/observation spaces to the canonical space, namely inverse LBS, can be defined with inverse transformation matrices.

\subsection{Overview}

The goal of \nickname{} is to train a generalizable Human NeRF model which can synthesize novel views and poses of 3D humans from a single image input.
For the input human image, we assume the calibrated camera parameters and the human region masks are known. 
We also assume the corresponding SMPL parameters $\{\bm{\theta}, \bm{\beta}\}$) are given.

The overall framework of \nickname{} is shown in Fig.~\ref{fig: overview}. 
The input is a single human image $\mathbf{I}^{o}$ and its corresponding camera parameters $\bm{P}^{o}$ and SMPL pose $\bm{\theta}^{o}$ and shape $\bm{\beta}^{o}$.
The output is the human rendering in the target camera view $\bm{P}^{t}$ with SMPL pose $\bm{\theta}^{t}$ and shape $\bm{\beta}^{t}$.
To render an image in the target space, we cast rays and sample points $\bm{x}^{t}$ along the ray.
$\bm{x}^{t}$ is transformed to the canonical space $\bm{x}^{c}$ through the inverse LBS.
We then query the bank of 3D-aware hierarchical features, \ie, global feature $\bm{f}_{global}$, point-level feature $\bm{f}_{point}$ and pixel-aligned feature $\bm{f}_{pixel}$, from their corresponding feature extraction modules.
To efficiently integrate features from the feature bank, we further apply a feature fusion transformer to get the fused features $\bm{f}_{trans}(\bm{x}^{c})$ for $\bm{x}^{c}$ as follows:
\begin{equation}
\begin{aligned}
\bm{f}_{trans}(\bm{x}^{c})=\text{Attn}(\bm{f}_{global}(\bm{x}^{c}), \bm{f}_{point}(\bm{x}^{c}), \bm{f}_{pixel}(\bm{x}^{c})).
\end{aligned}
\end{equation}
The point $\bm{x}^{c}$, along with the fused features $\bm{f}_{trans}(\bm{x}^{c})$, are fed into the NeRF decoder to predict the density $\sigma$ and RGB $\bm{c}$.
Finally, volume rendering is performed in the target space to render the pixel colors by integrating the density and RGB values of sampled 3D points along the rays in the target space.
In the following parts, we introduce details of the hierarchical feature extraction scheme and the feature fusion transformer.

\subsection{Hierarchical Feature Extraction}
The bank of 3D-aware hierarchical features comprises global, point-level and pixel-aligned features. 
With both global and fine-grained information learned from the bank of hierarchical features, \nickname{} enhances information observable from the input image and complements the information missing from the partial observation.

\noindent \textbf{Global Feature.}
Capturing the global structure and overall appearance is essential for recovering Human NeRF from partial observations. As shown in previous work~\cite{chen2018implicit_decoder, dupont2020equivariant, jang2021codenerf, OccupancyNetworks, DVR, Park_2019_CVPR, rematas2021sharf}, compressing the whole scene into a compact latent code $\bm{f}_{global}$ helps the encoding of such global information.
Therefore, as shown in Fig.~\ref{fig: overview}, we compress the input image into a compact latent code using a 2D encoder.
To efficiently decode the 3D representation from the compact latent code, we involve the tri-plane representation~\cite{Chan2021}, which plays an important role in missing information completion.
The compact latent code is first mapped to a 512-dimensional style vector through the mapping network~\cite{Karras2019stylegan2}.
The style vector is then fed into the style-based encoders~\cite{Karras2019stylegan2} to generate features which are further reshaped to the tri-plane representation to model Human NeRF in the canonical space.
Finally, points transformed to the canonical space $\bm{x}^{c}$ are projected to 3 planes through orthogonal projection to extract the 3D-aware global features $\bm{f}_{global}(\bm{x}^{c})$.

\noindent \textbf{Point-Level Feature.}
For single image Human NeRF, it is important to recover both global structure and local details from the input image, which can be bridged by an underlying explicit human model, \ie SMPL.
We first extract per-point features by projecting SMPL vertices to the 2D feature map of the input image, as shown in Fig.~\ref{fig: overview}.
In the single human image input setting, one problem with the above feature extraction process is that only half of the SMPL vertices are visible from the input view.
Therefore, to make the point-level feature aware of occlusions, we only extract features for vertices visible from the current camera $\bm{P}^o$.
Then we perform inverse LBS to transform the posed vertices features to the canonical space, which are then voxelized to sparse 3D volume tensors and further processed by sparse 3D convolutions~\cite{spconv2022}.
From the encoded sparse 3D volume tensors, we can extract point-level features $\bm{f}_{point}(\bm{x}^{c})$ for point $\bm{x}^{c}$.
The point-level features are aware of the 3D human structure and local texture details, which are helpful to infer textural information of Human NeRF in a more detailed level.

\noindent \textbf{Pixel-Aligned Feature.} 
The point-level features extraction uses SMPL prior and spatial convolution for both global and local feature enhancement.
However, it may suffer from significant information loss due to the limited SMPL mesh resolution and the voxel resolution.
To compensate for the fine-grained local information missing problem, we further extract the pixel-aligned features by projecting 3D deformed points $\bm{x}^{c}$ into the input view.
As shown in Fig.~\ref{fig: overview}, we transform the deformed point $\bm{x}^{c}$ to $\bm{x}^{o} = \text{LBS}(\bm{x}^{c}; \bm{\theta}^{o}, \bm{\beta}^{o})$ in observation space through LBS and project it to the input view so that pixel-aligned features $\bm{f}_{pixel}(\bm{x}^{c})$ can be queried, which can be formulated as
\begin{equation}
\label{feature: 2d}
\begin{aligned}
\bm{f}_{pixel}(\bm{x}^{c}) = \Pi(\bm{W}(\mathbf{I}^{o}); \text{LBS}(\bm{x}^{c}; \bm{\theta}^{o}, \bm{\beta}^{o})), 
\end{aligned}
\end{equation}
where $\bm{W}$ is the 2D feature encoder, $\Pi(\cdot)$ denotes the 3D-to-2D projection operator.
When used in the multi-view input setting, the variance of pixel-aligned features from different views can indicate whether 3D points are near the 3D surface or not.
While in our single image setting, the pixel-aligned features can not encode such implicit information, especially when the 3D points $\bm{x}^{c}$ are far from the surface.
To avoid overfitting to the uninformative pixel-aligned features, we assign different weights\footnote{If the L2 norm distance is large than a threshold $0.05$, we directly set the alpha $\bm{\sigma}$ and color $\bm{c}$ of 3D point as a small value, \eg, $\bm{\sigma}=-80$ and $\bm{0}$.} to pixel-aligned features according to the distance between the corresponding 3D deformed point $\bm{x}^{c}$ and its nearest SMPL vertex.

\subsection{Feature Fusion Transformer}
The above hierarchical features effectively encode different levels of texture and 3D structural information. However, it is not trivial to fuse these features.
Intuitively, for the observable parts, we should rely more on the pixel-aligned feature for the finest level of texture recovery. While for the invisible parts, the global features and the point-level features, where more coherent 3D-aware information are encoded, should contribute more.
To model such complex feature relations, we employ the self-attention module $\text{Attn}(\cdot)$~\cite{vaswani2017attention, devlin2018bert, dosovitskiy2020image} with three attention heads for effective feature fusion, \ie, 
\begin{equation}
\label{feature: transformer}
\begin{aligned}
\text{Attn}(Q,K,V) = \text{Softmax}(\frac{QK^{T}}{\sqrt{d}})V,
\end{aligned}
\end{equation}
where query $Q$, key $K$ and value $V$ are obtained by projecting the hierarchical features using MLPs. $d$ is the scaling coefficient.
The fused features are input into the NeRF decoder to predict the density $\bm{\sigma}$ and RGB $\bm{c}$ for the point $\bm{x}^t$.

\begin{table*}[t]
\caption{Performance (PSNR, SSIM and LPIPS) comparison among NHP, MPS-NeRF and our \nickname{} method on the THuman, RenderPeople, ZJU\_MoCap and HuMMan datasets. 
% $\uparrow$ means the larger is better; $\downarrow$ means the smaller is better.
}
\centering
\label{tab: main_redult}
% \resizebox{1\linewidth}{!}{
\small
\setlength{\tabcolsep}{1.2mm}{
\begin{tabular}{l|cccccc|cccccc}
\toprule
\multirow{3}*{Method} & \multicolumn{6}{c|}{\textit{THuman}} & \multicolumn{6}{c}{\textit{RenderPeople}} \\
\cline{2-13}
 & \multicolumn{3}{c}{Novel View} & \multicolumn{3}{c|}{Novel Pose} & \multicolumn{3}{c}{Novel View} & \multicolumn{3}{c}{Novel Pose} \\
% \cline{2-7}  
~ & PSNR$\uparrow$ & SSIM$\uparrow$ & LPIPS$\downarrow$ & PSNR$\uparrow$ & SSIM$\uparrow$ & LPIPS$\downarrow$ & PSNR$\uparrow$ & SSIM$\uparrow$ & LPIPS$\downarrow$ & PSNR$\uparrow$ & SSIM$\uparrow$ & LPIPS$\downarrow$\\
\hline\hline
PixelNeRF~\cite{yu2021pixelnerf} & 16.51 & 0.65 & 0.35 & - & - & - & - & - & - & - & - & - \\
NHP~\cite{kwon2021neural} & 22.53 & 0.88 & 0.17 & 20.25 & 0.86 & 0.19 & 20.59 & 0.81 & 0.22 & 19.60 & 0.77 & 0.25 \\
MPS-NeRF~\cite{gao2022mps} & 21.72 & 0.87 & 0.18 & 21.68 & 0.87 & 0.18 & 20.72 & 0.81 & 0.24 & 20.19 & 0.80 & 0.25 \\
\textbf{\nickname{} (Ours)} & \textbf{24.66} & \textbf{0.91} & \textbf{0.10} & \textbf{24.26} & \textbf{0.91} & \textbf{0.11} & \textbf{22.88} & \textbf{0.88} & \textbf{0.14} & \textbf{21.98} & \textbf{0.86} & \textbf{0.15} \\
\midrule
\multirow{3}*{Method} & \multicolumn{6}{c|}{\textit{ZJU\_MoCap}} & \multicolumn{6}{c}{\textit{HuMMan}} \\
\cline{2-13}
 & \multicolumn{3}{c}{Novel View} & \multicolumn{3}{c|}{Novel Pose} & \multicolumn{3}{c}{Novel View} & \multicolumn{3}{c}{Novel Pose} \\
~ & PSNR$\uparrow$ & SSIM$\uparrow$ & LPIPS$\downarrow$ & PSNR$\uparrow$ & SSIM$\uparrow$ & LPIPS$\downarrow$ & PSNR$\uparrow$ & SSIM$\uparrow$ & LPIPS$\downarrow$ & PSNR$\uparrow$ & SSIM$\uparrow$ & LPIPS$\downarrow$\\
\hline\hline
NHP~\cite{kwon2021neural} & 21.66 & 0.87 & 0.17 & 21.57 & 0.87 & 0.17 & 18.99 & 0.84 & 0.18 & 18.32 & 0.83 & 0.18 \\
MPS-NeRF~\cite{gao2022mps} & 21.86 & 0.87 & 0.17 & 21.60 & 0.87 & 0.17 & 17.44 & 0.82 & 0.19 & 17.43 & 0.82 & 0.19 \\
\textbf{\nickname{} (Ours)} & \textbf{22.87} & \textbf{0.89} & \textbf{0.12} & \textbf{22.38} & \textbf{0.89} & \textbf{0.12} & \textbf{20.83} & \textbf{0.89} & \textbf{0.12} & \textbf{20.43} & \textbf{0.88} & \textbf{0.11} \\
\bottomrule
\end{tabular}}
% \vspace{-3mm}
\end{table*}

\subsection{Training Details}
\nickname{} contains five trainable modules, which are three hierarchical feature extraction modules, the feature fusion transformer and the NeRF decoder, which are trained in an end-to-end manner. 
During training, for the same actor, we randomly sample image pairs from target and input views. By inputting the input view images to the above described process, we aim to reconstruct the actor in the target view. Four loss functions are used to supervise the training.

\noindent \textbf{Photometric Loss.} 
Given the ground truth target image $C(\mathbf{r})$  and predicted image $\hat{C}(\mathbf{r})$, we apply the photometric loss as follows:
\begin{equation}
\label{loss: color}
\begin{aligned}
\mathcal{L}_{color} = \frac{1}{|\mathcal{R}|}\sum_{\mathbf{r}\in \mathcal{R}} ||\hat{C}(\mathbf{r}) - C(\mathbf{r})||_2^2,
\end{aligned}
\end{equation}
where $\mathcal{R}$ denotes the set of rays, and $|\mathcal{R}|$ is the number of rays in $\mathcal{R}$.

\noindent \textbf{Mask Loss.} 
We also leverage the human region masks for Human NeRF optimization. The mask loss is defined as:
\begin{equation}
\label{loss: mask}
\begin{aligned}
\mathcal{L}_{mask} = \frac{1}{|\mathcal{R}|}\sum_{\mathbf{r}\in \mathcal{R}} ||\hat{M}(\mathbf{r}) - M(\mathbf{r})||_2^2,
\end{aligned}
\end{equation}
where $\hat{M}(\mathbf{r})$ is the accumulated volume density and $M(\mathbf{r})$ is the ground truth binary mask label.

\noindent \textbf{SSIM Loss.} 
We further employ SSIM to ensure the structural similarity between ground truth and synthesized images, \ie,  
\begin{equation}
\label{loss: ssim}
\begin{aligned}
\mathcal{L}_{SSIM} = \text{SSIM}(\hat{C}(\mathbf{r}), C(\mathbf{r})).
\end{aligned}
\end{equation}

\noindent \textbf{LPIPS Loss.} 
The perceptual loss LPIPS is also utilized to ensure the quality of rendered image, \ie,
\begin{equation}
\label{loss: lpips}
\begin{aligned}
\mathcal{L}_{LPIPS} = \text{LPIPS}(\hat{C}(\mathbf{r}), C(\mathbf{r})).
\end{aligned}
\end{equation}

\noindent In summary, the overall loss function contains four components, \ie, 
\begin{equation}
\label{loss: overall}
\begin{aligned}
\mathcal{L} = \mathcal{L}_{color} + \lambda_{1}\mathcal{L}_{mask} + \lambda_{2}\mathcal{L}_{SSIM} + \lambda_{3}\mathcal{L}_{LPIPS},
\end{aligned}
\end{equation}
where $\lambda$'s are the loss weights. 
Empirically, we select $\lambda_1=0.1$, $\lambda_2 = \lambda_3 = 0.01$ to ensure the same magnitude for each loss term.

\section{Experiments}
\label{sec:exp}

\begin{figure*}[t]
    % \vspace{-5mm}
    % \setlength{\abovecaptionskip}{0.1cm}
    \centering
    \includegraphics[width=18cm]{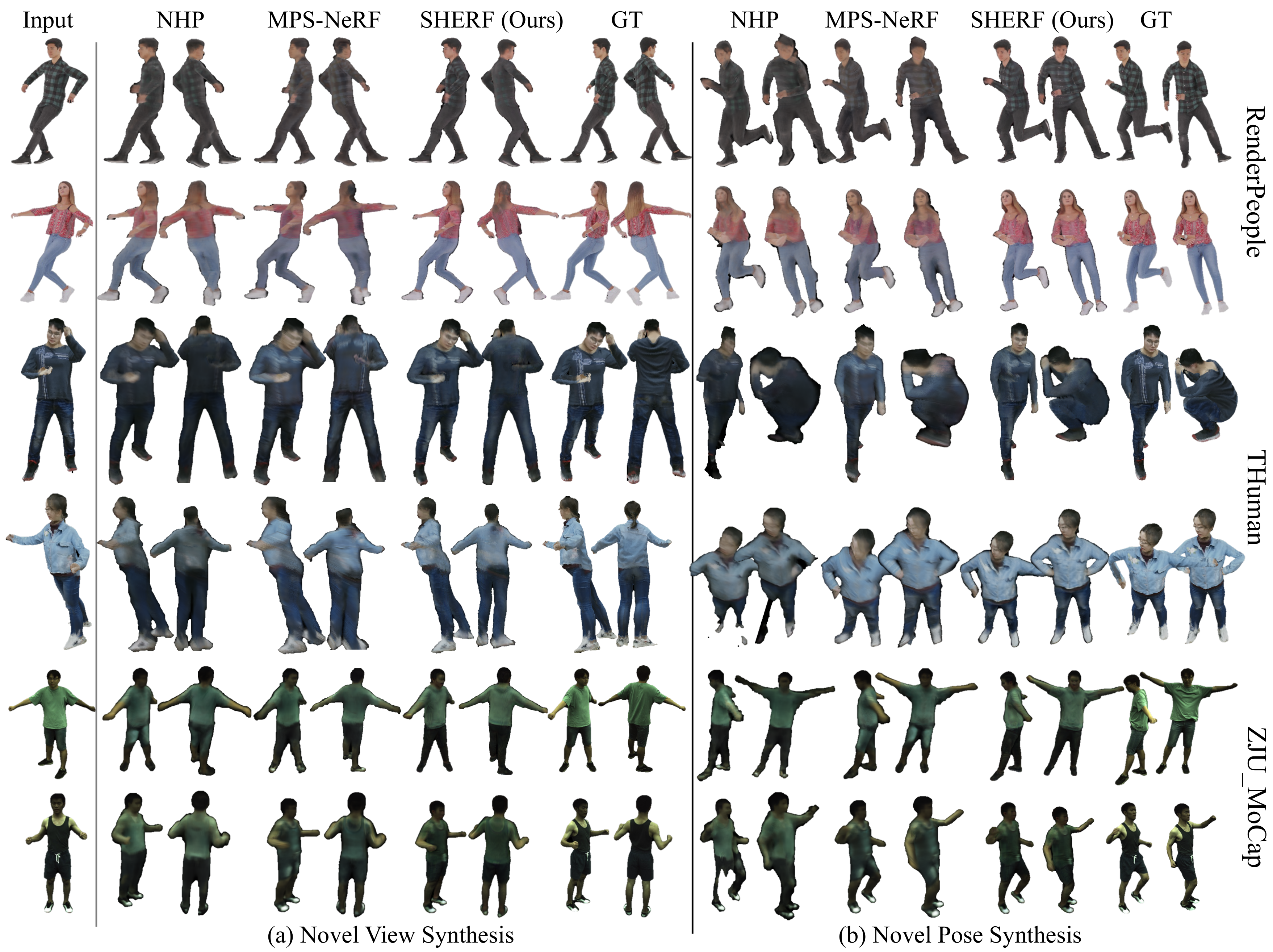}
    \caption{Novel view and novel pose synthesis results produced by NHP, MPS-NeRF and \nickname{} on RenderPeople, THuman and ZJU\_MoCap. Zoom in for the best view.} 
\label{fig: SHERF_main_figure_vis}
% \vspace{-3mm}
\end{figure*}

\subsection{Experimental Setup}

\noindent \textbf{Datasets.}
Four large-scale human datasets are used for evaluation, \ie, THuman~\cite{zheng2019deephuman}, RenderPeople~\cite{renderpeople}, ZJU\_MoCap~\cite{peng2021neural} and HuMMan~\cite{cai2022humman}.
For ZJU\_MoCap, 9 subjects are split into 6 for training and 3 for testing. 
For each training and test subject, 100 frames are sampled for training or evaluation.
For THuman, we randomly select 90 subjects as the training set and 10 subjects for testing.
For each subject, we randomly sample 20 frames for training or evaluation.
For RenderPeople, we randomly sample 450 subjects as the training set and 30 subjects for testing. 
For each subject, we randomly sample 10 frames for training or evaluation.
For HuMMan, we use 317 sequences as the training set and 22 sequences for testing, following the official data split  (HuMMan-Recon) for human reconstruction studies.
For each sequence, we randomly sample 17 frames for training or evaluation.

\noindent \textbf{Comparison Methods.}
To the best of our knowledge, we are the first to study the setting of single-image generalizable and animatable Human NeRF. We adapt two state-of-the-art generalizable Human NeRF methods designed for multi-view settings, \ie, NHP~\cite{kwon2021neural} and MPS-NeRF~\cite{gao2022mps} to our setting. For fair evaluation, we also compare with PixelNeRF~\cite{yu2021pixelnerf}, which is a generalizable NeRF reconstruction methods using single images as inputs. PixelNeRF is not specifically designed for human. Therefore, we only evaluate it for novel view synthesis.

\noindent \textbf{Implementation Details.}
Our 2D Encoder adopts a pretrained ResNet18 backbone to extract 1D vector $\bm{f}\in\mathcal{R}^{512}$ for global feature extraction, and feature maps $\bm{f}\in\mathcal{R}^{64\times256\times256}$ for point-level and pixel-aligned feature extraction.
To preserve more low-level information, we further perform positional encoding to the RGB values and append them to 2D feature maps to form feature maps $\bm{f}\in\mathcal{R}^{96\times256\times256}$.
The Mapping Network and Style-Based Encoder are adopted from EG3D~\cite{Chan2021}. 
Four layers of sparse convolutions~\cite{spconv2022} are used to extract 96-dimensional point-level features.
The queried global feature, point-level feature and pixel-aligned feature are concatenated and projected to 32 channels before fed into the feature fusion transformer.
The feature fusion transformer contains one self-attention layer with three heads.
NeRF decoder is the same as that used in~\cite{gao2022mps}. 
For LBS and inverse LBS of SMPL, we use the transformation or inverse transformation matrix of the nearest SMPL vertex.

\noindent \textbf{Evaluation Metrics.}
To quantitatively evaluate the quality of rendered novel view and novel pose images, we report the peak signal-to-noise ratio (PSNR) ~\cite{sara2019image}, structural similarity index (SSIM)~\cite{wang2004image} and Learned Perceptual Image Patch Similarity (LPIPS)~\cite{zhang2018unreasonable}.
Instead of computing the metrics for the whole image, we follow previous Human NeRF methods~\cite{neuralbody, gao2022mps} to project the 3D human bounding box to each camera plane to obtain the bounding box mask and report these metrics based on the masked area.

\begin{table*}[t]
\centering
\makebox[0pt][c]{\parbox{1.0\textwidth}{
\begin{minipage}[c]{0.75\hsize}
    \centering
    \caption{
        Ablation study on THuman. The left side shows different design components that are ablated on.
    }
    \vspace{-5pt}
    \label{tab:ablation}
    \small{
    \addtolength{\tabcolsep}{-3.pt}
    \begin{tabular}{cccc|cccccc}
        \toprule
        \multirow{2}*{\makecell{Global \\ Feature}} & \multirow{2}*{\makecell{Point-Level \\ Feature}} & \multirow{2}*{\makecell{Pixel-Aligned \\ Feature}} & \multirow{2}*{\makecell{Feature \\ Fusion}} & \multicolumn{3}{c}{Novel View} & \multicolumn{3}{c}{Novel Pose} \\
        \cmidrule{5-10}
        ~ & ~ & ~ & ~ & PSNR$\uparrow$ & SSIM$\uparrow$ & LPIPS$\downarrow$ & PSNR$\uparrow$ & SSIM$\uparrow$ & LPIPS$\downarrow$ \\
        \midrule
        $\checkmark$ & & & & 22.38 & 0.89 & 0.14 & 22.35 & 0.89 & 0.14 \\
        $\checkmark$ & $\checkmark$ & & & 23.26 & 0.89 & 0.13 & 23.03 & 0.89 & 0.14 \\
        $\checkmark$ & & $\checkmark$ & & 23.72 & 0.90 & 0.12 & 23.65 & 0.90 & 0.12 \\
         & $\checkmark$ & $\checkmark$ & & 24.08 & 0.90 & 0.12 & 23.73 & 0.90 & 0.12 \\
        $\checkmark$ & $\checkmark$ & $\checkmark$ & & 24.44 & \textbf{0.91} & 0.11 & 24.08 & \textbf{0.91} & \textbf{0.11} \\
        $\checkmark$ & $\checkmark$ & $\checkmark$ & $\checkmark$ & \textbf{24.66} & \textbf{0.91} & \textbf{0.10} & \textbf{24.26} & \textbf{0.91} & \textbf{0.11} \\
        \bottomrule
    \end{tabular}}
\end{minipage}
\hfill
\begin{minipage}[c]{0.22\hsize}
    \begin{center}
        \includegraphics[width=1.0\linewidth]{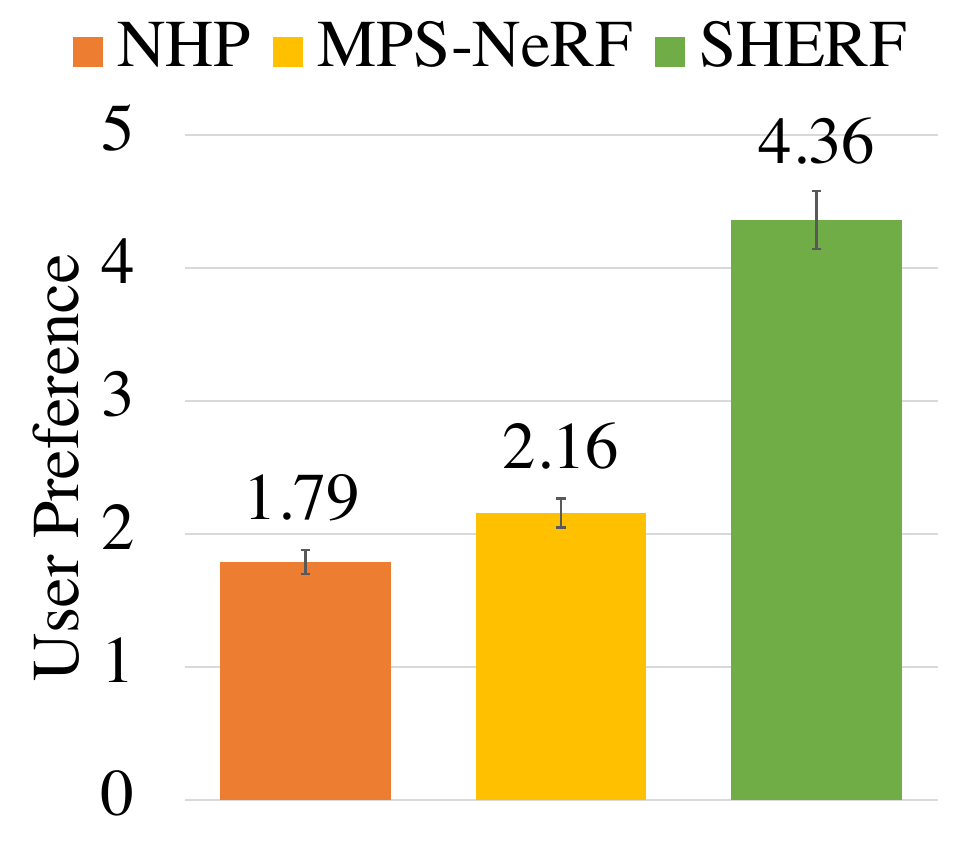}
    \end{center}
    \vspace{-16pt}
    \captionof{figure}{User preference scores on rendering results.}
    \label{fig:user_study}
\end{minipage}
}}
\end{table*}

\subsection{Quantitative Results}
As shown in Tab.~\ref{tab: main_redult}, \nickname{} significantly outperforms NHP and MPS-NeRF in all evaluation metrics on all four datasets.
NHP and MPS-NeRF, as the SOTA generalizable Human NeRF methods in multi-view human image input setting, achieve reasonable performance in the novel view synthesis task.
However, both NHP and MPS-NeRF focus on the local feature extraction and lack the ability to complement information missing from partial inputs, explaining their poor performance.
They also fail to have good performance in the novel pose synthesis task, especially for NHP, which models neural radiance field in the target space.
In contrast, \nickname{} achieves the best performance on both novel view synthesis and novel pose synthesis tasks.
In addition to these three metrics, we also perform a user study and report human's preference scores on rendered images. As shown in Fig.~\ref{fig:user_study}, \nickname{} has a clear advantage over two baseline methods.

\subsection{Qualitative Results}
We show the rendering images of our \nickname{} and two baseline methods in Fig.~\ref{fig: SHERF_main_figure_vis}.
NHP and MPS-NeRF produce reasonable RGB renderings in novel views, but they fail to recover details, \eg, face details and cloth patterns. 
Thanks to the bank of hierarchical features, our \nickname{} successfully recovers face details and cloth patterns by enhancing the information from the input 2D observation and complementing the information missing from the input image.
For example, \nickname{} renders the 3D human cloth with the same patterns as the input image and the 3D human back cloth with reasonable colors which are not observable from the input image.
In novel pose synthesis, NHP produces distorted rendering results as it models the neural radiance filed in the target space.
Compared with MPS-NeRF, \nickname{} shows better rendering results in novel pose synthesis.

\subsection{Ablation Study}
\label{sec:exp}
\begin{figure}[t]
    % \vspace{-2mm}
    % \setlength{\abovecaptionskip}{0cm}
    \centering
    \includegraphics[width=8.5cm]{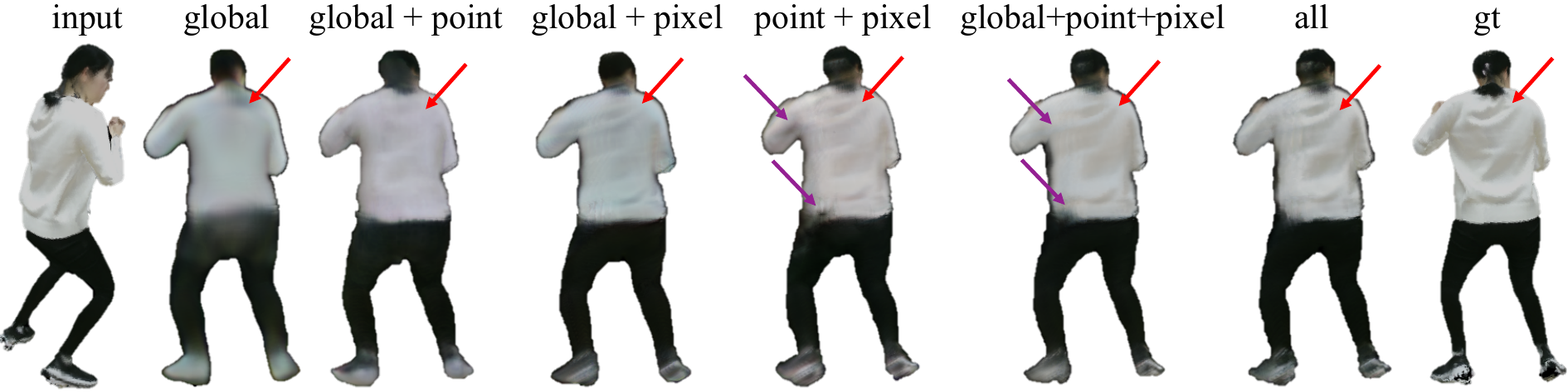}
    \caption{Qualitative results of ablation study on THuman. Refer to red arrows to see the cloth color and wrinkle difference, and purple arrows to see the erasion of black artifacts.} 
\label{fig:ablation}
% \vspace{-3mm}
\end{figure}

To validate the effectiveness of the proposed hierarchical feature extraction components and feature fusion transformer, we subsequently add different components and evaluate the performance on the THuman dataset.
The results are reported on Tab.~\ref{tab:ablation} and one visualization example is shown in Fig.~\ref{fig:ablation}.
% where red arrows are used to mark the cloth color and wrinkle difference.
Given a single image input, only using global features can render images with reasonable RGB colors on test subjects.
Adding point-level features or pixel-aligned features can further improve the image quality.
% and render images with colors more similar as the ground truth, while adding pixel-aligned features can render images with more fine-grained details, \eg, the cloth wrinkle in Fig.~\ref{fig:ablation}. 
Qualitatively, pixel-aligned features preserves more fine-grained details, \eg, the cloth wrinkle in Fig.~\ref{fig:ablation}.
Combining both point-Level and pixel-aligned features with global features can further improve the performance and render images with correct colors and fine-grained details, erasing the small artifacts when only point-level and pixel-aligned features are used (see purple arrows in Fig.~\ref{fig:ablation}).
Finally, using the feature fusion transformer, we further improve the reconstruction quality.
% For example, more realistic cloth wrinkle details can be observed in Fig.~\ref{fig:ablation}.

\subsection{Further Analysis}

\noindent \textbf{Discussion on Training Protocols.}
Previous generalizable Human NeRF methods~\cite{gao2022mps} carefully pick input camera views and fix them during training.
Switching to a more challenging setting of single image input, a straightforward way is to extend this training setting by only using the front view as input.
% The results on THuman dataset are reported on Tab.~\ref{tab: front_view_redult}. 
We evaluate the model trained with this setting with different settings.
As shown in Tab.~\ref{tab: front_view_redult}, when evaluated in either setting, \nickname{} leads the performance. Especially, model shows high rendering result when evaluated with the front view input setting.
However, in most real-world scenarios, human images are captured from random camera viewing angles, challenging the above front view input training setting.
Therefore, we further evaluate the above model using random view inputs.
As shown in the right half of Tab.~\ref{tab: front_view_redult}, the performance is much worse than front view inputs. It is also worse than the model trained with free view inputs in Tab.~\ref{tab: main_redult}.
% The above shows our free view input training setting is a more reasonable choice for real-world scenarios.
The conclusion is that instead of a fixed viewing angle input, our free view input training setting is a more reasonable choice for real-world scenarios.
\begin{table}[h]
% \vspace{-3mm}
% \setlength{\abovecaptionskip}{0cm}
\caption{Performance (PSNR) comparison among NHP, MPS-NeRF and \nickname{} trained with front view input and evaluated with different settings on THuman dataset.
}
\centering
\label{tab: front_view_redult}
% \resizebox{1\linewidth}{!}{
\small
\setlength{\tabcolsep}{0.4mm}{
\begin{tabular}{lcccc}
\toprule
\multirow{3}*{Method} & \multicolumn{2}{c}{Front View Input} & \multicolumn{2}{c}{Free View Input} \\
\cmidrule{2-5} 
~ & Novel View & Novel Pose & Novel View & Novel Pose \\
\midrule
% \multicolumn{7}{c}{Front View Input}\\
\midrule
NHP~\cite{kwon2021neural} & 24.00 & 19.75 & 20.59 & 19.17\\
MPS-NeRF~\cite{gao2022mps} & 23.29 & 23.15 & 21.56 & 21.46 \\
\textbf{\nickname{} (Ours)} & \textbf{24.63} & \textbf{24.05} & \textbf{22.60} & \textbf{22.36} \\
\bottomrule
\end{tabular}}
% \vspace{-2mm}
\end{table}

\begin{figure}[h]
    % \vspace{-3mm}
    \centering
    \includegraphics[width=8cm]{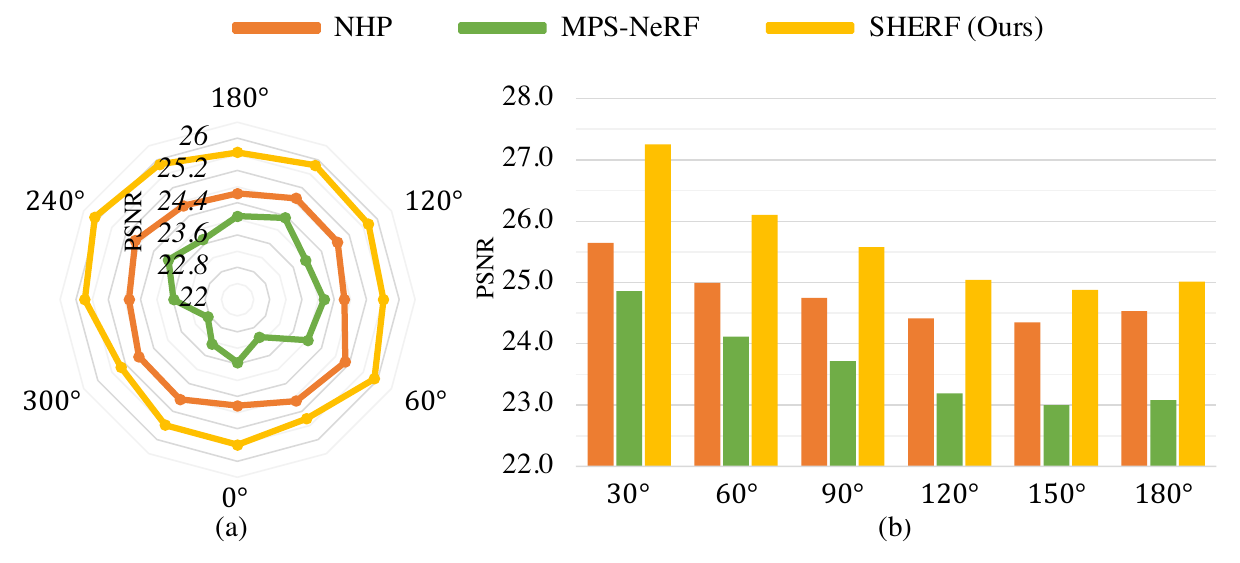}
    % \setlength{\abovecaptionskip}{0cm}
    % \setlength{\belowcaptionskip}{0.1cm}
    % \vspace{-10pt}
    \caption{\textbf{Analysis on Input Viewing Angles.} (a) reports novel view synthesis PSNR with different viewing angles as inputs. (b) shows novel view synthesis PSNR with different viewing angle difference between targets and inputs.} 
\label{fig: analysis}
% \vspace{-3mm}
\end{figure}

\noindent \textbf{Analysis on Different Viewing Angles as Inputs.}
Further evaluations are performed to better understand how \nickname{} would perform given images with different viewing angles.
We evaluate \nickname{} and baseline methods with input images from 12 camera viewing angles, which are evenly sampled from [$0^\circ$, $360^\circ$].
For each input view, the target is to render the other 11 views, on which the mean PSNR is calculated and reported in Fig.~\ref{fig: analysis}(a).
We find that \nickname{} 1) consistently outperforms SOTA baseline methods on all viewing angle inputs, and 2) shows robust performance to different viewing angle inputs.

\noindent \textbf{Analysis on Viewing Angle Difference Between Target and Observation.} 
As shown in Fig.~\ref{fig: analysis}(b), we study the effect of viewing angle difference between targets and inputs on the novel view synthesis.
Two main trends can be found.
1) The smaller the viewing angle difference is, the easier for models to perform novel view synthesis.
2) Across all input settings, \nickname{} consistently outperforms baseline methods.
By refining the indicators, we provide more detailed evaluation and gain more insight into the models.

\begin{figure}[h]
    % \vspace{-3mm}
    \centering
    \includegraphics[width=8cm]{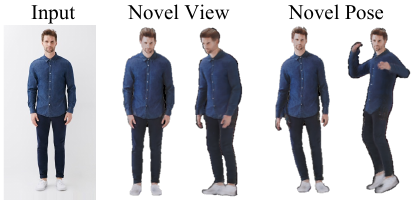}
    % \setlength{\abovecaptionskip}{0cm}
    % \setlength{\belowcaptionskip}{0.1cm}
    % \vspace{-10pt}
    \caption{Generalization ability of SHERF on in-the-wild DeepFashion images.} 
\label{fig: generalizability_deepfashion}
% \vspace{-3mm}
\end{figure}

\noindent \textbf{Generalizability Analysis.}
To further study the generalization ability, we directly inference NHP, MPS-NeRF and \nickname{} models, which are pre-trained on THuman, on ZJU\_Mocap test sets.
As shown in Tab.~\ref{tab: cross_validation_redult}, without additional fine-tuning, \nickname{} achieves better performance than NHP and MPS-NeRF.
\nickname{} even shows comparable performance with NHP and MPS-NeRF that are trained on ZJU\_MoCap.
To further show the generalizability on in-the-wild images, we evaluate \nickname{} on DeepFashion~\cite{liu2016deepfashion} images with SMPL and camera estimated from a 2D image using CLIFF~\cite{li2022cliff} and show novel view and pose synthesis visualization results in Fig.~\ref{fig: generalizability_deepfashion}.

\begin{table}[t]
% \vspace{-2mm}
% \setlength{\abovecaptionskip}{0cm}
\caption{Generalization ability comparison by cross-validating NHP, MPS-NeRF and \nickname{} trained with THuman on ZJU-MoCap.
% $\uparrow$ means the larger is better; $\downarrow$ means the smaller is better.
}
\centering
\label{tab: cross_validation_redult}
% \resizebox{1\linewidth}{!}{
\small
\setlength{\tabcolsep}{0.3mm}{
\begin{tabular}{lcccccc}
\toprule
\multirow{2}*{Method} & \multicolumn{3}{c}{Novel View} & \multicolumn{3}{c}{Novel Pose} \\
\cmidrule{2-7} 
~ & PSNR$\uparrow$ & SSIM$\uparrow$ & LPIPS$\downarrow$ & PSNR$\uparrow$ & SSIM$\uparrow$ & LPIPS$\downarrow$ \\
\midrule
NHP~\cite{kwon2021neural} & \textbf{22.07} & 0.88 & 0.16 & 20.70 & 0.86 & 0.18 \\
MPS-NeRF~\cite{gao2022mps} & 20.36 & 0.85 & 0.17 & 19.96 & 0.85 & 0.17 \\
\textbf{\nickname{} (Ours)} & 21.87 & \textbf{0.89} & \textbf{0.11} & \textbf{21.50} & \textbf{0.88} & \textbf{0.12} \\
\bottomrule
\end{tabular}
}
% \vspace{-3mm}
\end{table}

\begin{figure}[h]
    % \vspace{-3mm}
    \centering
    \includegraphics[width=6cm]{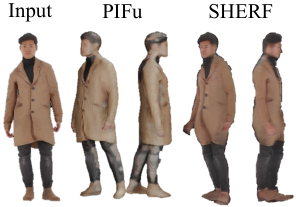}
    \caption{Visualization results comparison between PIFu and our \nickname{}.} 
\label{fig: result_pifu}
% \vspace{-3mm}
\end{figure}

\noindent \textbf{Comparison with PIFu.}
For PIFu series, as PIFuHD cannot predict 3D human with textures, we opt to compare with PIFu, which predicts textured mesh from single images.
We evaluate PIFu on RenderPeople using the official models and compare with \nickname\ in terms of visualization results.
Thanks to the volume rendering in NeRF and direct supervision on images, \nickname{} can generate textures with higher fidelity and better details than PIFu as shown in Fig.~\ref{fig: result_pifu}. 
Moreover, 3D human reconstructed by PIFu cannot be animated, while \nickname{} can naturally generalize to novel poses.

\noindent \textbf{Comparison with modulation and adaptive normalization.}
Even though global, point-level and pixel-aligned features contain different levels of semantics, they share the same shape, and is better suited for a transformer fusion module, where the learnable attention map allows sufficient flexibility.
We compare our transformer fusion module with other two classic feature fusion methods, \ie, AdaIN and SIREN. We report their PSNR in novel view/pose synthesis in Tab.~\ref{tab: result_feature_fusion}, which shows our transformer fusion is the most effective one.

\begin{table}[h]
% \vspace{-2mm}
% \setlength{\abovecaptionskip}{0cm}
\caption{Performance (PSNR, SSIM and LPIPS) comparison among modulation, adaptive normalization and our Transformer fusion on the RenderPeople dataset.
}
\centering
\label{tab: result_feature_fusion}
\small
\setlength{\tabcolsep}{0.3mm}{
\begin{tabular}{lcccccc}
\toprule
\multirow{2}*{Method} & \multicolumn{3}{c}{Novel View} & \multicolumn{3}{c}{Novel Pose} \\
\cmidrule{2-7} 
~ & PSNR$\uparrow$ & SSIM$\uparrow$ & LPIPS$\downarrow$ & PSNR$\uparrow$ & SSIM$\uparrow$ & LPIPS$\downarrow$ \\
\midrule
AdaIN & 22.38 & 0.87 & 0.16 & 21.65 & 0.85 & 0.17 \\
SIREN & 21.62 & 0.85 & 0.18 & 21.09 & 0.84 & 0.19 \\
\textbf{Transformer (Ours)} & \textbf{22.88} & \textbf{0.88} & \textbf{0.14} & \textbf{21.98}  & \textbf{0.86} & \textbf{0.15} \\
\bottomrule
\end{tabular}
}
\end{table}

\noindent \textbf{Runtime Analysis.}
One big advantage of generalizable Human NeRF is that the reconstruction happens in only one forward pass. The inference time is essential for down-stream applications. Therefore, we also report and compare the inference frames per second (FPS) of \nickname{} and baseline methods in Tab.~\ref{tab: runtime_analysis}.
\begin{table}[h]
% \vspace{-2mm}
% \setlength{\abovecaptionskip}{0cm}
\caption{Runtime comparison between NHP, MPS-NeRF and \nickname{}. ``FPS'' represents inference frames per second, the higher the better.}
% \vspace{-10pt}
\centering
\label{tab: runtime_analysis}
\small
\begin{tabular}{lccc}
\toprule
Method & NHP & MPS-NeRF & \nickname{} \\
\midrule
FPS & 0.15 & 0.60 & \textbf{1.33} \\
\bottomrule
\end{tabular}
\end{table}

\section{Discussion and Conclusion}
To conclude, we propose \textbf{\nickname{}}, the first generalizable human NeRF model that recovers animatable 3D humans from single human image inputs.
To render high-fidelity 3D humans, \nickname{} proposes to learn both global and local details from the bank of 3D-aware hierarchical features comprising global features, point-level features, and pixel-aligned features. 
By using a feature fusion transformer, \nickname{} successfully enhances the information from the 2D observation and complements the information missing from the input image. 
On four large-scale human datasets, \nickname{} achieves state-of-the-art performance and renders high-fidelity images in both novel views and poses.

\vspace{1.75mm}
\noindent \textbf{Limitations:}
1) There still exists visible artifacts in target renderings when some body parts are occluded in the observation space. 
A better feature presentation like occlusion-aware features may be explored to solve this issue. 
2) How to complement the information missing from single image input remains a challenging problem.
\nickname{} starts from the reconstruction view and can only render deterministic results when predicting novel views.
One potential direction is to investigate the use of conditional generative models to diversely generate higher quality novel views.

\vspace{1.75mm}
\noindent \textbf{Potential Negative Societal Impacts:} 
\nickname{} can be misused to create fake images or videos of real humans and cause negative social impacts.

\section{Acknowledgment}
This study is supported by the Ministry of Education, Singapore, under its MOE AcRF Tier 2 (MOE-T2EP20221-0012), NTU NAP, and under the RIE2020 Industry Alignment Fund – Industry Collaboration Projects (IAF-ICP) Funding Initiative, as well as cash and in-kind contribution from the industry partner(s).

% \newpage

{\small
\bibliographystyle{ieee_fullname}
\bibliography{egbib}
}

% \clearpage
\appendix
\section{Implementation Details}

\subsection{More Implementation Details of \nickname{}}
SHERF model is trained with images from different actors at the same time. For example, when sampling data pairs from THuman (90 subjects $\times$ 20 poses $\times$ 24 views), we randomly sample one input and one target image from the same subject. 
To render the target image during training and evaluation, we randomly sample an input image from given camera views and sample 48 points for each the ray belong to the human region bound box part at the target space.
During the optimization, we use the Adam~\cite{kingma2014adam} optimizer. 
We set the initial learning rate as $2\times 10^{-3}$ and decay the learning rate by a factor of $0.5$ for every epoch.
The maximum iteration number is set as 5 epochs.

\subsection{Novel Pose Synthesis of NHP}
As discussed in the main paper, there lacks a clear framework to synthesize novel poses in NHP~\cite{kwon2021neural} as it models the neural radiance field in the canonical space. 
In this work, we synthesis novel pose results of NHP by using the Linear Blend Skinning of SMPL algorithm. 
Specifically, we transform the 3D sampled points from the target space to observation space and query the corresponding features.
Then queried features, along with the coordinates of 3D sampled points and ray directions in the target space, are fed into the NeRF decoder to predict density $\bm{\sigma}$ and RGB $\bm{c}$ values.

\section{Analysis on SMPL and Camera Parameters Estimated from a 2D Input Image.}
\begin{figure}[h]
    % \vspace{-3mm}
    \centering
    \includegraphics[width=8cm]{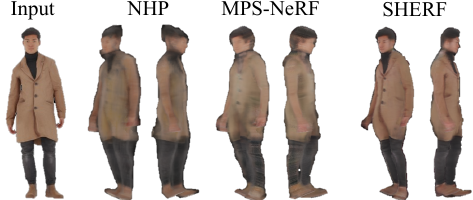}
    \setlength{\abovecaptionskip}{0.1cm}
    \caption{Visualization results with SMPL and camera parameters estimated from a 2D input image among NHP, MPS-NeRF and our \nickname{} method on the RenderPeople dataset.} 
\label{fig: result_smpl_camera_cliff}
% \vspace{-3mm}
\end{figure}

Current human NeRF methods, including multi-view images or monocular video settings, rely on accurate SMPL parameters.
In our experiments, we also use accurate SMPL and camera parameters provided by the datasets. 
Recently, single-view SMPL estimation methods have made great progress and are reliable.
To verify the effectiveness of our proposed \nickname{} in real-world scenarios with only one single 2D image available and no accurate SMPL and camera parameters available, we use SMPL and camera parameters predicted by CLIFF~\cite{li2022cliff} to evaluate the performance on the RenderPeople dataset. 
As shown in the Tab.~\ref{tab: result_smpl_camera_cliff} and Fig.~\ref{fig: result_smpl_camera_cliff}, SHERF produces plausible results and surpasses baseline methods.

\begin{table}[t]
% \vspace{-2mm}
% \setlength{\abovecaptionskip}{0cm}
\caption{Performance (PSNR, SSIM and LPIPS) comparison with SMPL and camera parameters estimated from a 2D input image among NHP, MPS-NeRF and our \nickname{} method on the RenderPeople dataset.}
\centering
\label{tab: result_smpl_camera_cliff}
\small
\setlength{\tabcolsep}{0.3mm}{
\begin{tabular}{lcccccc}
\toprule
\multirow{2}*{Method} & \multicolumn{3}{c}{Novel View} & \multicolumn{3}{c}{Novel Pose} \\
\cmidrule{2-7} 
~ & PSNR$\uparrow$ & SSIM$\uparrow$ & LPIPS$\downarrow$ & PSNR$\uparrow$ & SSIM$\uparrow$ & LPIPS$\downarrow$ \\
\midrule
NHP~\cite{kwon2021neural} & 18.04 & 0.72 & 0.31 & 17.59 & 0.70 & 0.33 \\
MPS-NeRF~\cite{gao2022mps} & 17.81 & 0.74 & 0.30 & 17.33 & 0.71 & 0.32 \\
\textbf{\nickname{} (Ours)} & \textbf{19.64} & \textbf{0.79} & \textbf{0.22} & \textbf{19.22} & \textbf{0.78} & \textbf{0.24} \\
\bottomrule
\end{tabular}
}
\end{table}

\section{More Qualitative Results}

\subsection{Models Trained with Free View Inputs}
\begin{figure*}[t]
    \centering
    \includegraphics[width=15.5cm]{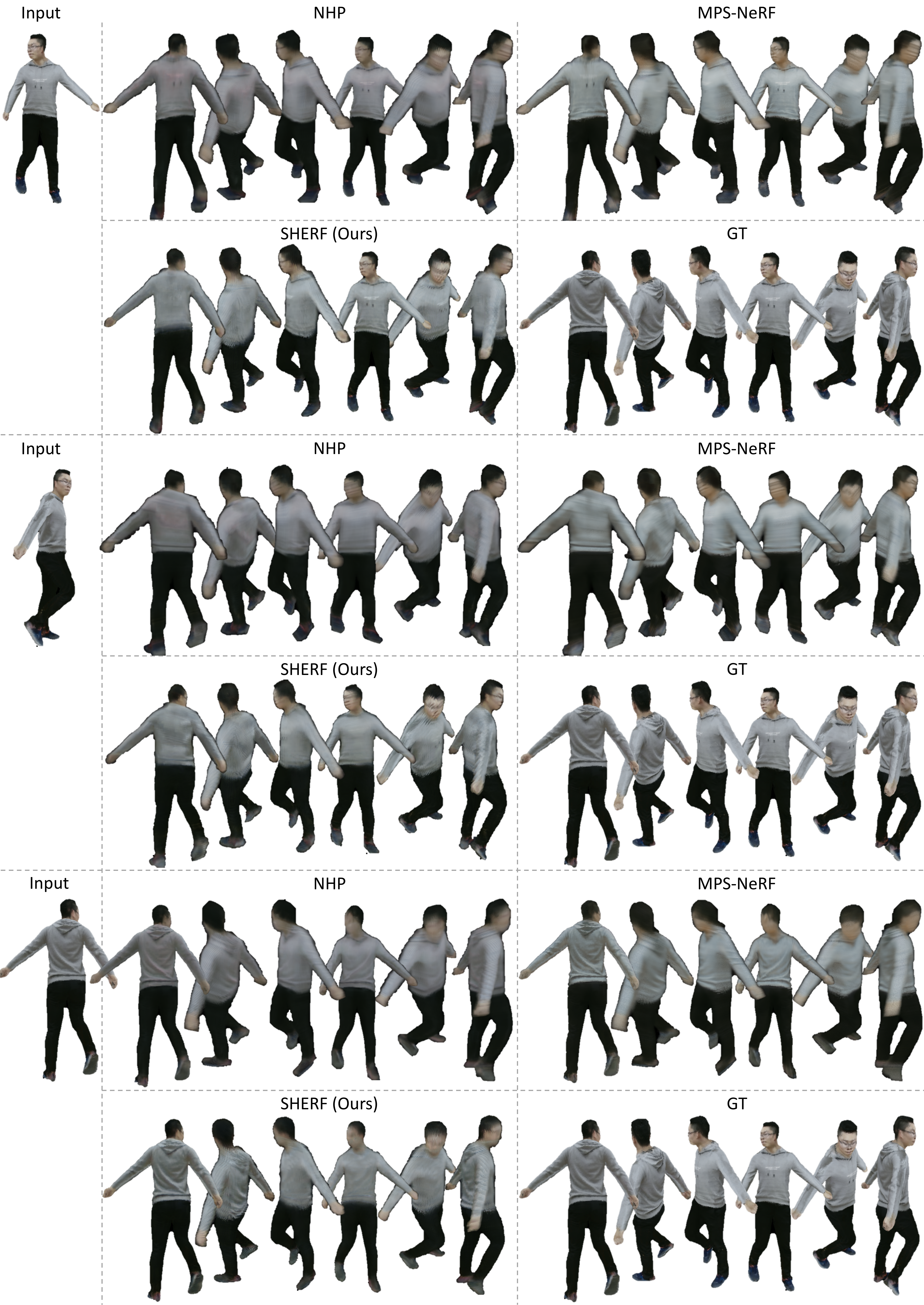}
    \caption{More qualitative results with different viewing angles as inputs on test subjects of THuman. } 
    \label{app: free_view_1}
\end{figure*}

\begin{figure*}[t]
    \centering
    \includegraphics[width=15.5cm]{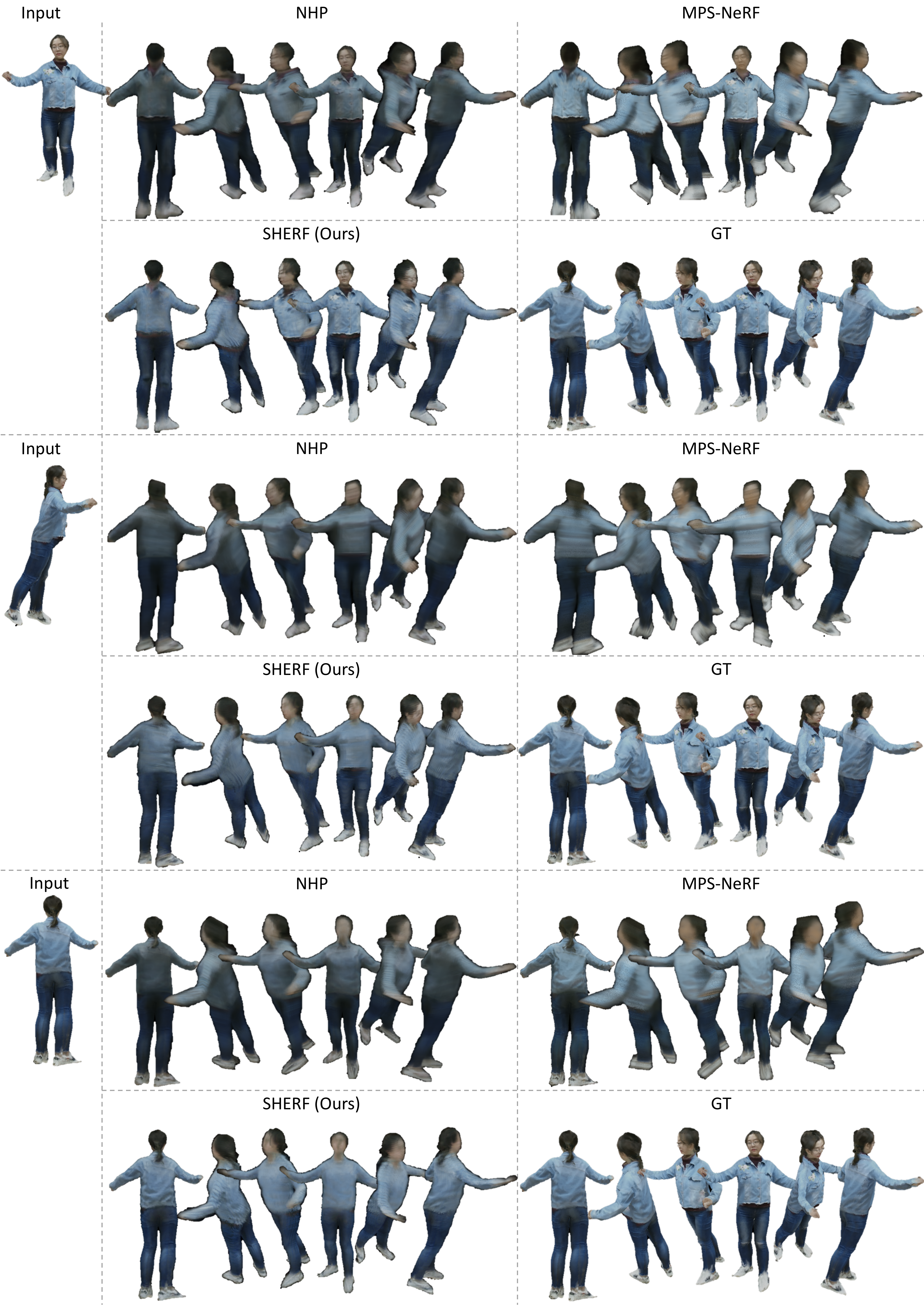}
    \caption{More qualitative results with different viewing angles as inputs on test subjects of THuman.}
    \label{app: free_view_4}
\end{figure*}

More qualitative results with different viewing angles as inputs on test subjects of THuman are shown in Fig.~\ref{app: free_view_1} - Fig.~\ref{app: free_view_4}. 
The models are trained with free viewing angles as inputs on training subjects of THuman.
Two main trends can be observed.
1) NHP~\cite{kwon2021neural} tends to render images with smoothed effects in face and cloth, failing to produce realistic image details.
MPS-NeRF~\cite{gao2022mps} can somehow produce image details, but still suffers from recovering face details.
Thanks to the bank of hierarchical features, our \nickname{} can render more realistic images with details in face and cloth when compared with NHP and MPS-NeRF.
2) When given the front viewing angle input, NHP and MPS-NeRF overfit to the cloth patterns of the front view input image when synthesizing the back view output image while our \nickname{} can learn to synthesize images with more acceptable results.
3) When given the back viewing angle input, NHP and MPS-NeRF fails to render images with reasonable face details especially for the front viewing angle output, while our \nickname{} can generate results with acceptable image quality.
For more qualitative results in RenderPeople data set, please refer to our demo video. 

\subsection{Models Trained with Front View Inputs}
\begin{figure*}[t]
    \centering
    \includegraphics[width=15.5cm]{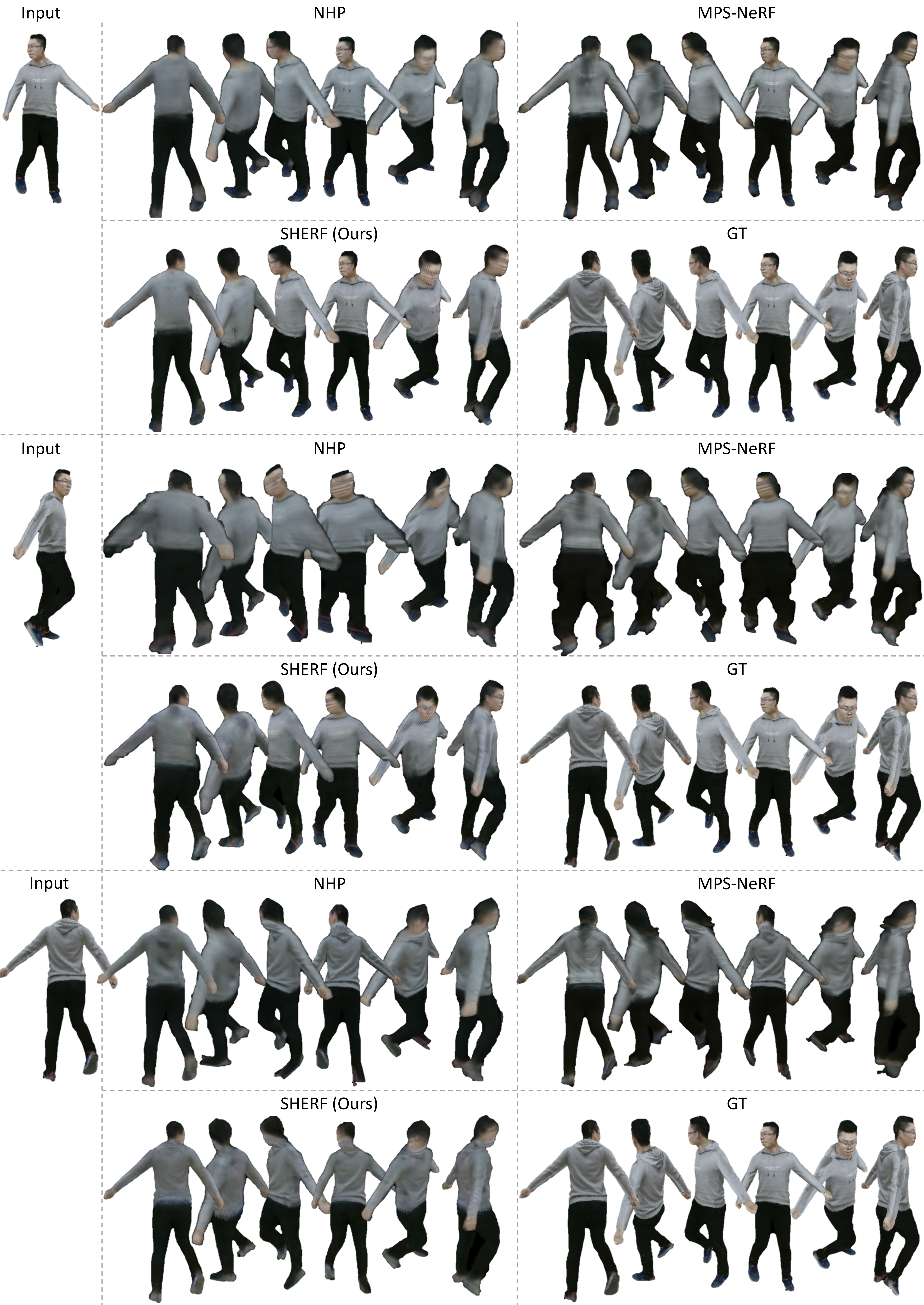}
    \caption{More qualitative results with different viewing angles as inputs on models trained only with front view inputs of THuman.} 
    \label{app: fix_front_view_1}
\end{figure*}

\begin{figure*}[t]
    \centering
    \includegraphics[width=15.5cm]{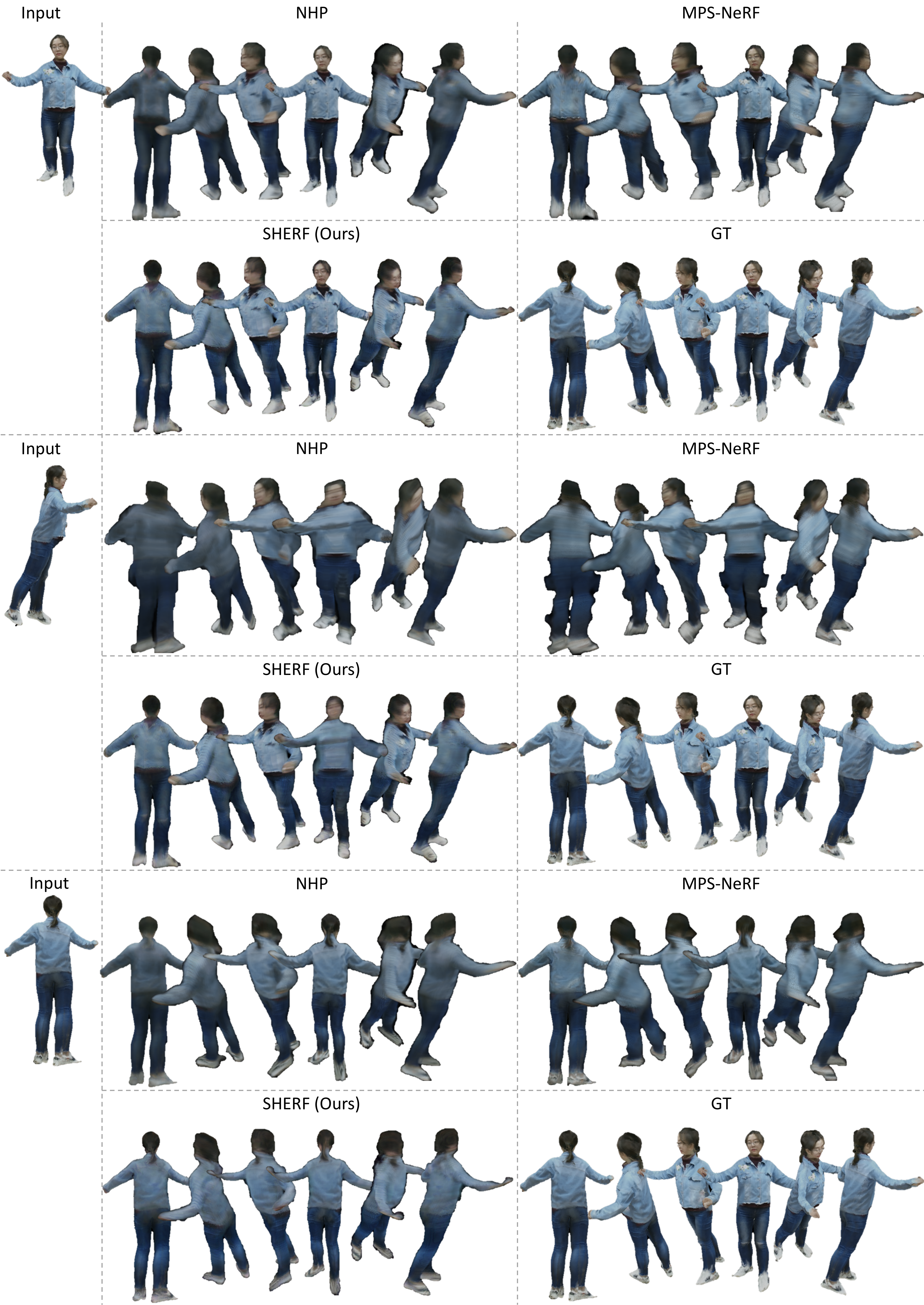}
    \caption{More qualitative results with different viewing angles as inputs on models trained only with front view inputs of THuman.} 
    \label{app: fix_front_view_4}
\end{figure*}

In the analysis part, we show that models trained with front view inputs are not suitable for the real-world scenarios where human images are captured individually from a random camera viewing angle.
To further support our claim, we show qualitative results with different viewing angles as inputs on models trained only with front view inputs of THuman.
As shown in Fig.~\ref{app: fix_front_view_1} - Fig.~\ref{app: fix_front_view_4}, although all three methods can produce good results with front view inputs, the image quality degrades significantly when other free viewing angle inputs are provided.
For example, when given the back viewing angle input, almost no reasonable results can be produced.
Even in the front view input setting, \nickname{} still produces better results when compared with two SOTA baseline methods.

\end{document}